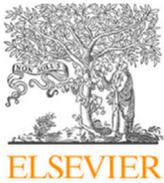
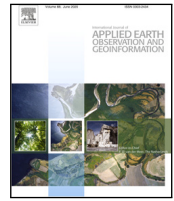

# PolyU-BPCoMa: A dataset and benchmark towards mobile colorized mapping using a backpack multisensorial system

Wenzhong Shi[1], Pengxin Chen[1,*], Muyang Wang, Sheng Bao, Haodong Xiang, Yue Yu, Daping Yang

*Department of Land Surveying and Geo-Informatics, The Hong Kong Polytechnic University, Hong Kong 999077, China*
*Smart Cities Research Institute, The Hong Kong Polytechnic University, Hong Kong 999077, China*

ABSTRACT

Constructing colorized point clouds from mobile laser scanning and images is a fundamental work in surveying and mapping. It is also an essential prerequisite for building digital twins for smart cities. However, existing public datasets are either in relatively small scales or lack accurate geometrical and color ground truth. This paper documents a multisensorial dataset named PolyU-BPCoMa which is distinctively positioned towards mobile colorized mapping. The dataset incorporates resources of 3D LiDAR, spherical imaging, GNSS and IMU on a backpack platform. Color checker boards are pasted in each surveyed area as targets and ground truth data are collected by an advanced terrestrial laser scanner (TLS). 3D geometrical and color information can be recovered in the colorized point clouds produced by the backpack system and the TLS, respectively. Accordingly, we provide an opportunity to benchmark the mapping and colorization accuracy simultaneously for a mobile multisensorial system. The dataset is approximately 800 GB in size covering both indoor and outdoor environments. The dataset and development kits are available at https://github.com/chenpengxin/PolyU-BPCoMa.git.

## 1. Introduction

In the research fields of robotics, surveying and mobile mapping, publicly available datasets are playing significant roles for developing and evaluating algorithms. Over the past decade, many known datasets such as KITTI (Geiger et al., 2012), Oxford RobotCar (Maddern et al., 2017) and ISPRS MIMAP (Wen et al., 2020) have been released to provide a standard algorithm benchmark and keep benefiting the community.

Among the existing datasets that relate to mobile mapping, there are a variety of sensor platforms targeting different applications and road passing conditions. Multiple sensors are carried by vehicles, unmanned robots or humans. Vehicular platforms can only operate on outdoor roads; unmanned robots can take sensors to work in more flexible environments such as indoor rooms; and human-carried platforms are featured with finer manipulation of sensor movements so that they can be applied in mapping places that would otherwise be difficult to access with a mobile robot. In the human-carried category, hand-held and backpack systems are two popular fashions of sensor integration. Due to the limitations on volume and weight, the sensor resources and battery life are an obvious bottleneck which discourages a hand-held system from working in challenging scenarios. In the pursuit of fast and flexible surveying in challenging locations, the backpack multisensorial systems emerged to fill such a gap (Gong et al., 2021).

With the integration of laser scanner, panoramic camera, GNSS and IMU, we aim to provide a backpack multisensorial dataset that is well-rounded for mobile colorized mapping in various environments. For places with abundant geometrical features, precise localization and mapping can be achieved by point cloud registration. If geometrical changes are not sufficient for scan matching, the panoramic images can be utilized for matching textural features. In outdoor environments, a GNSS receiver provides us with real-time kinematic positioning (RTK) which is essential for the drift control during long-distance SLAM tasks. Considering the places that lack both geometrical and textural information, the fusion of RTK-GNSS and IMU can still ensure reliable localization. The most challenging environments should be those places where not only GNSS is denied, but also geometrical and textual features are degenerated, such as city tunnels. Under such a circumstance, pedestrian dead reckoning (PDF) can also be utilized to facilitate sensor fusion.

**Table 1**
Comparison of related datasets used in mobile localization and mapping research.

| Dataset | Year | Envrnmt | LiDAR | Camera | IMU | GNSS | Ground Truth | Platform |
|---|---|---|---|---|---|---|---|---|
| Darpa Urban | 2010 | outdoor | 12 SICK LMS291@75 Hz Velodyne HDL-64E@15 Hz | Point Grey: 4 × 376 × 240@10 Hz 1 × 752 × 480@22.8 Hz | Applanix POS-LV 220 | | GPS + INS | Car |
| Ford Campus | 2011 | outdoor | Velodyne HDL-64E@10 Hz 2 Riegl LMS-Q120 | Ladybug 3: 6 × 1600 × 600@8 Hz | Applanix POS-LV 420 | | GPS + INS | |
| KITTI | 2013 | outdoor | Velodyne HDL-64E@10 Hz | 4 Point Grey (2 gray + 2 RGB): 4 × 1392 × 512@10 Hz | OXTS RT3003 | | GPS + INS | |
| Oxford RoboCar | 2017 | outdoor | 2 SICK LMS151@50 Hz SICK LD-MRS@12.5 Hz | BumbleBee XB3: 2 × 1280 × 960@16 Hz 3 Grasshoper2: 3 × 1024 × 1024@11.1 Hz | NovAtel SPAN-CPT | | GPS + INS + VO | |
| Oxford Radar RobotCar | 2020 | outdoor | 2 HDL-32E@20 Hz 2 SICK LMS151@50 Hz | BumbleBee XB3: 2 × 1280 × 960@16 Hz 3 Grasshoper2: 3 × 1024 × 1024@11.1 Hz | NovAtel SPAN-CPT | | GPS + INS + VO | |
| Rawseeds | 2009 | indoor+ outdoor | 2 URG-04LX@10 Hz SICK LMS291@75 Hz SICK LMS200@75 Hz | Trinocular: 3 × 640 × 480@30 Hz Monocular: 640 × 480@30 Hz Omnidirectional: 640 × 640@15 Hz | Xsense MTi | | indoor: visual markers +2D ICP outdoor: GPS | Wheeled robot |
| New College | 2009 | outdoor | 2 SICK LMS 291@75 Hz | BumbleBee: 2 × 512 × 384@20 Hz Ladybug 2: 5 × 384 × 512@3 Hz | Segway base gyro@28 Hz | CSI Series @5 Hz | N/A | |
| NCLT | 2015 | indoor+ outdoor | 2 UTM-30LX@10/40 Hz Velodyne HDL-32E@10 Hz | Ladybug 3: 6 × 1600 × 1200@5 Hz | Microstrain GX3-45: acc/gyro/orien @100 Hz | NovAtel DL-4P | GPS + SLAM | |
| SubT-Tunnel | 2020 | tunnel | Ouster OS1-64@10 Hz | Carnegie Multisense: color@10 Hz, depth@2 Hz FLIR Boson: 640 × 512@10 Hz | Microstrian GX5-25: acc/gyro/orien @500 Hz | N/A | professional surveyors | |
| EuRoc MAV | 2016 | indoor | N/A | 2 MT9V034: 2 × 752 × 480@20 Hz | ADIS: acc/gyro @200 Hz | N/A | laser track+ motion capture | UAV |
| Zurich MAV | 2017 | outdoor | N/A | Gopro: 1920 × 1080@30 Hz | acc/gyro@10 Hz | Fotokite onboard GPS | aerial photogrammetry | |
| NTU VIRAL | 2021 | indoor+ outdoor | 2 Ouster OS1-16@10 Hz | 2 uEye 1221 LE: 2 × 752 × 480@10 Hz | VectorNav VN100: acc/gyro/orien @385 Hz | N/A | laser tracker | |
| USVInLand | 2021 | river surface | LeiShen LS-16@10 Hz | Mynteye S2110: 640 × 480@20 Hz | LPMS-IG1@50 Hz | U-Blox ZEDF9@5 Hz | GPS + INS | USV |
| PennCOSYVIO | 2017 | indoor+ outdoor | N/A | 3 Gopro: 3 × 1920 × 1080@30 Hz 2 MT9V03d: 2 × 752 × 480@20 Hz | ADIS: acc/gyro@200 Hz 2 Tango: acc@128 Hz, gyro@100 Hz | N/A | visual tags | hand-held |
| TUM VI | 2018 | indoor+ outdoor | N/A | IDS stereo: 2 × 1024 × 1024@20 Hz | BMI160: acc/gyro@200 Hz | N/A | motion capture | |
| Newer College | 2020 | outdoor | Ouster OS1-64@10 Hz | D435i: 2 × 848 × 480@30 Hz | camera embedded: acc/gyro@650 Hz | N/A | ICP | |
| Google Cartographer | 2016 | indoor | 2 Velocyne VLP-16@20 Hz | N/A | acc/gyro@250 Hz | N/A | N/A | Backpack |
| ISPRS MIMAP | 2020 | indoor | Velodyne VLP-32C@20 Hz | N/A | N/A | N/A | TLS | |
| Our Dataset | 2021 | indoor+ outdoor | Velodyne VLP-32C@20 Hz VLP-16 Hi-Res@20 Hz | Ladybug 5P: 8192 × 4096@30 Hz | Xsens MTi-300: acc/gyro/orien @400 Hz | Emlid Reach M2 | TLS + color checker | |

In addition to the comprehensiveness of sensor types, the volume of our dataset is also considerable. The total size of our dataset is approximately 800 GB. Two laser scanners (32 + 16 laser beams) rotate orthogonally at 20 Hz to collect point clouds in a full field of view. Ladybug 5P provides stitched panoramic images at an impressive resolution of 8192 × 4096@30 Hz which, to the best knowledge of the authors, offers the highest quality in spherical 360° imaging and accuracy among the existing related datasets (see Table 1). Our dataset covers both indoor and outdoor environments including some special places such as underground tunnels and staircases. For each data sequence, the ground-truth colorized point cloud is produced by a high-end terrestrial laser scanner (TLS).

Different sensor carriers and targeting applications yield different benchmarks. Many vehicular datasets use a manner of GPS/INS fusion to provide ground-truth trajectories. Although GPS can ensure global consistency with the absolute world coordinate system, such ground-truth trajectories lack local accuracy arising from the loss or reacquisition of satellite signals. This also explains why the KITTI odometry benchmark does not use length scales less than 100 m for the evaluation.[2] The ISPRS MIMAP dataset pays more attention on the mapping accuracy evaluation, so it uses the point cloud data generated from a TLS as ground truth. However, its target-free benchmark has two limitations: (1) the manual alignment of MLS and TLS point clouds will introduce random manual errors; (2) the absolute mapping error is affected by the voxel size for the subsample process. In contrast, the proposed dataset introduces color checkers as the targets for the accuracy benchmarking. We provide an opportunity to evaluate the mapping and colorization accuracy simultaneously for a mobile multisensorial system. Geometrical error is defined as the average distance from a vertex point of a color checker observed in the MLS point cloud to its counterpart point observed in the TLS point cloud, while the colorization error is evaluated by comparing the hue differences of color checkers in the HSL color space.

The main contributions of our work are as follows:

(1) We contribute a dataset that is distinctively positioned towards mobile colorized mapping. The dataset comprises 800 GB sensor data from 3D LiDAR, spherical imaging, GNSS-RTK/IMU on a backpack platform, and TLS point cloud data as ground truth.
(2) We provide an opportunity to benchmark the mapping and colorization accuracy simultaneously for a mobile multisensorial system.
(3) We provide a development kit for using this dataset and a baseline result for performance comparison.

Finally, we hope that this dataset will be useful to the vision and surveying community, and be helpful for developing algorithms for SLAM, mobile mapping and colorization.

## 2. Related work

As the proposed dataset aims to provide multisensorial data for mobile colorized mapping, we only review the prior datasets that are on mobile platforms and can be used for localization or mapping. According to the platform differences, we divide the related datasets into three groups comprising vehicular, unmanned, and human-carried datasets. Table 1 gives a comparison of some representative datasets in these three groups.

### 2.1. Vehicular dataset

Over the past decade, many vehicular multisensorial datasets have sprung up. The Darpa Urban (Huang et al., 2010) is one of the first large-scale datasets targeting localization and mapping. It records environment by thirteen 2D/3D laser scanners and five cameras through which users can develop their SLAM algorithms. The Ford Campus (Pandey et al., 2011) precisely registers 3D laser scans onto omnidirectional images, thereby adding visual information to individual laser scans. Similar to the Ford dataset, the KITTI dataset (Geiger et al., 2012) uses a 64-beam laser scanner and stereo cameras to collect geometrical and visual information of surroundings. Based on these data, KITTI also provides challenging computer vision benchmarks concerning lidar/visual odometry, depth completion, stereo/scene flow evaluation and object detection.

The above datasets generate ground-truth trajectories using a fashion of GPS/INS fusion which can guarantee the global consistency with the absolute world coordinate system but lack local accuracy arising from the loss or reacquisition of satellite signals. To address this issue, the Oxford RobotCar dataset (Maddern et al., 2017), one of the longest (1000 km) autonomous driving datasets, uses visual odometry to smoothen local poses. As an extension to Maddern et al. (2017), the Oxford Radar RobotCar dataset (Barnes et al., 2020) adds a milimeter–wave radar sensor and replaces the previous planar scanner with two 32-beam spherical counterparts.

Utilizing the point clouds and street views produced by vehicular datasets, many downstream datasets for semantic segmentation, object classification and detection have drawn increasing interest and attention in recent years. Representative datasets include TUM-Campus (Gehrung et al., 2017), SemanticKITTI (Behley et al., 2019), nuScenes (Caesar et al., 2020), and Audi (Geyer et al., 2020) datasets.

### 2.2. Unmanned dataset

Unlike vehicular datasets that mainly target the autonomous driving, unmanned datasets focus more on mobile robotics research. Prior work has contributed valuable datasets to the community with different sensor carriers such as wheeled robot, Unmanned Aerial Vehicle (UAV) and Unmanned Surface Vehicle (USV).

The ground-truth trajectory or point cloud map is key information for benchmarking self-localization and mapping algorithms. Both the New College (Smith et al., 2009) and the NCLT (Carlevaris-Bianco et al., 2016) datasets use a Segway robot to log data, but the difference is that the former does not provide ground truth while the latter collects ground truth by RTK-GNSS and SLAM. The Zurich MAV (Majdik et al., 2017) dataset takes GPS readings as the initial position of its recorded images and performs an accurate photogrammetric 3D reconstruction using the Pix4D software to produce ground truth. In the USVInLand (Cheng et al., 2021) dataset, an unmanned boat is driven in inland waterways to collect information of various modalities including LiDAR, radar, vision, and GPS/INS. It provides a benchmark for SLAM, stereo matching and water segmentation. Its ground truth is generated by GPS/INS fusion and manual annotation.

Obtaining ground truth in GNSS-denied environment is difficult. The Rawseeds dataset (Ceriani et al., 2009) equips its robot with six black-and-white markers so that the ground truth is gathered by recovering the relative positions of the markers. The SubT-Tunnel (Rogers et al., 2020) places twenty artifacts in the robot's traverse, and these artifacts' positions are surveyed by professional surveyors to benchmark the mobile mapping accuracy. The EuRoc MAV (Burri et al., 2016) and the NTU VIRAL (Nguyen et al., 2022) datasets mount a prism on their drones, so that millimeter-level ground-truth positions can be acquired by tracking the prism using a laser tracker. However, continuous line of sight between the prism and laser tracker must be maintained all through data collection. This limits the size of experimental area as well as the scale of datasets.

### 2.3. Human-carried dataset

Human-carried platforms are designed for mapping places that would otherwise be difficult to access with a mobile robot. Here we distinguish our dataset from the related human-carried datasets in two aspect: data source and benchmarks.

**(1) Data source.** PennCOSYVIO (Pfrommer et al., 2017) and TUM VI (Schubert et al., 2018) are two hand-held datasets. They focus on evaluating visual inertial odometry, so they only integrate cameras and IMUs in their datasets. Both LiDAR and GNSS data are not available. By contrast, Google Cartographer (Hess et al., 2016) and ISPRS MIMAP (Wen et al., 2020) are two backpack datasets and they pay more attention to the mapping accuracy rather than the odometry result. The Google Cartographer dataset collect two 16-beam laser scanners and a 250 Hz IMU measurements without any visual information or ground truth data. The backpack used in the MIMAP dataset is equipped with two laser scanners and four fish-eye cameras, but thus far only three indoor traverses with only one 32-beam LiDAR's data has been available in its SLAM dataset. The Newer College dataset (Ramezani et al., 2020) can provide LiDAR/Visual/Inertial measurement, but its camera is in low resolution which might not viable for point cloud colorization.

---

[2] http://www.cvlibs.net/datasets/kitti/eval_odometry.php

Our dataset provides the most comprehensive data sources in the listed human-carried category. Two laser scanners can ensure a full sphere FOV. Acceleration, angular speed and orientation are measured at the frequency of 400 Hz. RTK-GNSS data is provided for geo-referencing as well as odometry benchmarking. Most impressively, our dataset collects panoramic images at 8192 × 4096@30 Hz (over 1 billion RGB pixels per second), which greatly contributes to colorized mapping, modeling and many downstream applications such as semantic segmentation and object detection. In addition, 91 stations of terrestrial point clouds are provided. They can be utilized for benchmarking not only the mobile colorized mapping, but also terrestrial point cloud registration.

**(2) Benchmarks.** The PennCOSYVIO and TUM VI datasets use visual tags and motion capture to collect ground-truth trajectories, respectively. The *absolute trajectory error* (ATE) and the *Relative Pose Error* are used in these two datasets for benchmarking visual odometry performance. The Newer College dataset generates ground-truth trajectories of centimeter-level accuracy by registering individual mobile laser sweeps to a terrestrial point cloud. In the ISPRS MIMAP dataset, the mapping accuracy benchmark can be summarized as three steps[3]: MLS/TLS clouds downsampling, manual alignment and point-to-point error computation.

By contrast, our benchmark focuses on evaluating both geometry and colorization accuracy. The geometry accuracy is evaluated by comparing the geometrical relationships between MLS and TLS clouds, while the colorization benchmark is partly inspired by the prior work (Julin et al., 2020). Although different metrics are used for these two benchmarks, the experiments can be conducted using just one sort of targets: the color checker. Our dataset, to the best knowledge of the authors, is the only public dataset utilizing color checker boards to benchmark the mapping and colorization performance simultaneously for a mobile mapping system.

## 3. Hardware design

Fig. 1 shows the hardware assembly of the backpack multisensorial system.

### 3.1. Sensors

The sensors used in the PolyU-BPCoMa dataset include:

**(1) Velodyne VLP-32C:** The VLP-32C is a long-range 3D LiDAR sensor. The detection range is up to 200 m. Its 32-channel laser beams provide [+15°, −25°] vertical and 360° horizontal field of views (FOVs). It can collect 600k points per second in the single return mode. It has a 3 cm typical range accuracy.

**(2) Velodyne VLP-16 Hi-Res:** The VLP-16 Hi-Res is a higher resolution version of the popular VLP-16. Its 16-channel laser beams provide a [+10°, −10°] vertical FOV. The higher vertical resolution (1.33°) enables it to capture more details in 3D mapping applications. It has a 3 cm typical range accuracy, too.

**(3) FLIR Ladybug5P:** The Ladybug5P is a 360° spherical camera with six CMOS sensors which provide 30 Megapixels (2464 × 2048 × 6) image resolution with a 3.45 μm pixel size and a 12-bit analog-to-digital converter. The camera operates in the readout manner of global shutter. The six lenses enable our backpack to collect video from more than 90% of the full sphere. In this dataset, Ladybug5P directly outputs stitched panoramic images from the six lenses at 8192 × 4096@30 Hz FPS.

**(4) Xsens MTi-300:** The Xsens MTi-300 is a high-performing Attitude and Heading Reference System (AHRS) which features vibration-rejecting gyroscopes and supports optimized temperature calibration.

---

[3] http://mi3dmap.net/isprsDatatype1.jsp

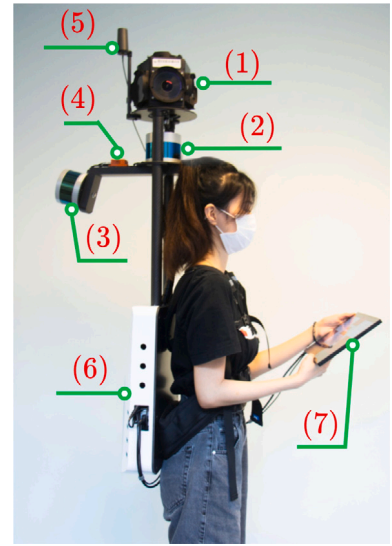

**Fig. 1.** The backpack multisensorial system used for data collection. Components include (1) omni-directional camera; (2) 32-beam laser scanner; (3) 16-beam laser scanner; (4) IMU; (5) GNSS receiver/antenna; (6) computer and battery case; (7) hand-held touch screen.

The standard full ranges are $450°/s$, 20 g and $\pm 8$ G for its gyroscope, accelerometer and magnetometer, respectively. The in-run bias stability is $10°/h$ for the gyroscope and 15 μg for the accelerometer. The noise density is $0.01°/s/\sqrt{Hz}$ for the gyroscope and $60 \mu g/\sqrt{Hz}$ for the accelerometer. For the magnetometer, its total RMS noise is 0.5 mG; non-linearity is 0.2%; and resolution is 0.25 mG.

**(5) Emlid Reach M2:** The Reach M2 is a compact GNSS receiver for positioning with centimeter accuracy. It outputs RTK fixed/float data at 5 Hz. The static accuracy is 4 mm + 0.5 ppm for horizontal positioning and 8 mm + 1 ppm for vertical positioning. The kinematic accuracy is 7 mm + 1 ppm for horizontal positioning and 14 mm + 1 ppm for vertical positioning. Utilizing the GNSS module, the colorized point cloud map can thus be geo-referenced.

### 3.2. System integration

On the backpack, the VLP-32C is placed horizontally while the VLP-16 Hi-Res is tilted 70°. The intention behind such a layout design is to maximize the benefit for both localization and mapping tasks. As the VLP-32C spins horizontally, it can hardly acquire ceiling and ground data in narrow environments such as corridors. To compensate, a VLP-16 Hi-Res is placed vertically to obtain ceiling and ground information. As a result, the VLP-32C can detect farther objects which are useful for localization while the VLP-16 Hi-Res works like a "push broom" to scan nearer objects for detailed mapping. Both scanners are spinning at 20 Hz for data collection.

Despite the accurate range measurement, LiDAR has a inherent weakness: its data do not contain RGB properties. Therefore, a panoramic imaging sensor is integrated into the system for color information. For the LiDAR or visual SLAM tasks, the odometry and mapping process will gradually accumulate drift errors. In light of this, GNSS is introduced to control the drift because the GNSS readings only contain global errors which will not drift with time or distance.

LiDAR, imaging and GNSS all rely on external observations. In short-term featureless GNSS-denied environments, these sensors would fail to localize themselves. The common practice is to add an inertial measurement unit to the system. By integrating an IMU that does not rely on external perceptions, the reliability of the data fusion can be increased. On the other hand, since the IMU measurements diverges quickly, other external observations, such as LiDAR/visual odometry and GNSS

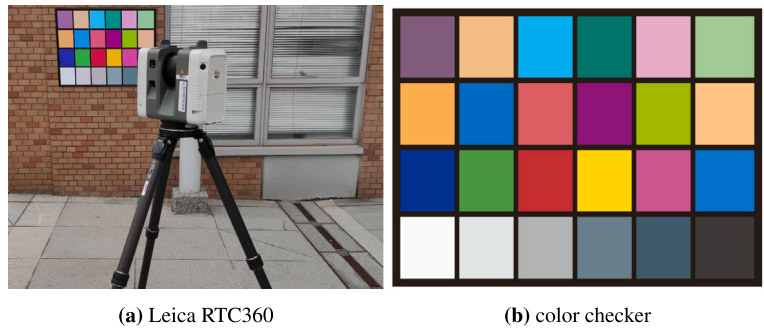

**(a)** Leica RTC360  **(b)** color checker

**Fig. 2.** Ground truth generation. (a) Leica RTC 360 and the backpack system separately scan the environments with multiple pasted color checkers. The colorized point clouds generated by the Leica RTC 360 are used as ground truth data. (b) a direct view of the color checker. (For interpretation of the references to color in this figure legend, the reader is referred to the web version of this article.)

positioning, are required to reset IMU's acceleration measurements periodically.

In addition to the mapping and navigation sensors, there are some other essential devices for the system integration. We use two 120k mAh Li-ion batteries to achieve over five hours system endurance. A LTE/Wi-Fi transmitter is added to the system, so that the GNSS receiver can connect to the Internet for RTK data. A network switching hub is used for the bridge connection from the two laser scanners to the onboard computer.

An onboard Intel NUC 11 i7 computer drives and synchronizes all sensors in the Robot Operating System (ROS)[4]. Since the Ladybug5P's linux SDK only supports Ubuntu 16.04 but the Wi-Fi 6 AX201 chip embedded in the NUC computer needs newer Linux kernel and firmware versions, we opt to use Ubuntu 16.04 with the Linux kernel of 5.13.0-generic and the Linux firmware package of 1.187.15 Focal Fossa.

## 4. Data collection and file formats

### 4.1. Backpack multisensorial data

The surveyed areas include outdoor terrace, indoor office, corridor, staircase and underground tunnel. In each area, we provide two data sequences with different traverse paths and walking speeds. The backpack system outputs $2\times20$ Hz $\times48$ scan lines, $30 \times 8$K panoramic images, 400 IMU measurements and 5 RTK-GNSS readings per second. The dataset provides the following data formats:

**(1) ROS bag.** A ROS bag file is packed with all sensor data collected during a traverse. LiDAR, GNSS, and IMU data are stored in the standard ROS format: namely the *sensor_msgs/PointCloud2, sensor_msgs/Imu, sensor_msgs/NavSatFix*. Image data are stored in a self-defined ROS format: *pointgrey_ladybug/ladybug_image*.

**(2) LiDAR data.** LiDAR scans in a single 360° sweep are extracted from the ROS bag file. The scan data from two laser scanners are spatially calibrated and temporally synchronized. To save space, a single scan is stored as a $N\times6$ float matrix into a binary file. $N$ denotes the number of laser points in a scan while the 6 means the 6 types of information of a point: namely the $x, y, z, intensity, ring, time$.

**(3) Image data.** While raw ladybug data are stored in a ROS bag file, single panoramic images are also extracted through post-processing in which six lenses of images are stitched together. The stitched panoramic image is then compressed to JPEG format with a 80% quality level.

**(4) GNSS/IMU data.** GNSS and IMU data are additionally stored in csv files. For a piece of GNSS reading, both WGS84 and local ENU coordinates are provided with a corresponding timestamp and a covariance matrix. The covariance matrix is used to represent the accuracy of the RTK positioning. For a IMU measurement, timestamp, orientation in quaternion, angular velocity, linear acceleration and their corresponding covariance matrices are provided.

### 4.2. Ground truth

Ground truth data are collected by a Leica RTC360 terrestrial laser scanner, as shown in Fig. 2(a). With 36 MP 3-camera system, the Leica RTC360 can capture 432 megapixels raw data for calibrated 360°×360° spherical HDR images. Its range accuracy is 1.0 mm + 10 ppm. The 3D point accuracy is 1.9 mm@10 m and 2.9 mm@20 m.

Multiple color checkers are pasted in the scanned environments. The size of color checkers is $80 \times 60$ cm and each of them contains 24 kinds of color squares (see Fig. 2(b)). Their 3D geometrical and color information will be recovered in the colorized point clouds produced by both the backpack system and the Leica TLS. The TLS outputs are utilized as ground truth data while the mobile mapping results from the backpack system are to be evaluated against the ground truth. Fig. 3 illustrates two sampled point cloud produced by the Leica TLS in both indoor and outdoor environments, together with the corresponding registration path. In the dataset, 91 stations of TLS data in E57 format with ground truth poses are provided.

Table 2 describes different segments of the dataset in terms of the location of the environments, number of GNSS/IMU measurements, number of panoramic images and pixels, number of scans and points acquired from the backpack system, number of stations and points acquired from the terrestrial scanners, etc.

## 5. Benchmarks

### 5.1. Benchmark pipeline

Our dataset provides an opportunity to benchmark the mapping and colorization accuracy simultaneously for a mobile multisensorial system. Fig. 4 gives an example result of mobile colorized mapping using the method described in Chen et al. (2021). The sub-figures Fig. 4(d)(e)(f) show the color checkers in the registered point cloud, colorized point cloud and the terrestrial point cloud, respectively. As the color checkers can be visually recognized, we can manually segment them out of these clouds and use them for accuracy evaluation. The evaluation pipeline are demonstrated in Fig. 5 where we employ six steps to quantitatively benchmark the geometrical error and colorization error.

For each data sequence, we set up multiple terrestrial stations to scan the environment. Therefore, the first step for the benchmark pipeline is the multiple-station TLS registration using Leica Cyclone REGISTER 360 software, after which multiple terrestrial scans are stitched together into one point cloud. In the step 2, we manually extract all color checkers in the stitched terrestrial point cloud. On the side of mobile mapping, dataset users will need to generate a colorized point cloud for each data sequence using their own SLAM and colorization algorithm, which corresponds to the steps 4 and 5 in the pipeline. Thereafter, they also need to extract the color checkers from the colorized point cloud manually (step 6). The green and red points produced from step 2 and 6 represent the vertex points of color checkers scanned by the TLS and the MLS separately, so they have one-to-one correspondences. Since these two groups of vertex points are in different coordinate systems, they are aligned by the ICP algorithm (Besl and McKay, 1992) in the step 3. Finally in the step 7, the geometrical error and the colorization error are computed.

---
[4] https://www.ros.org/

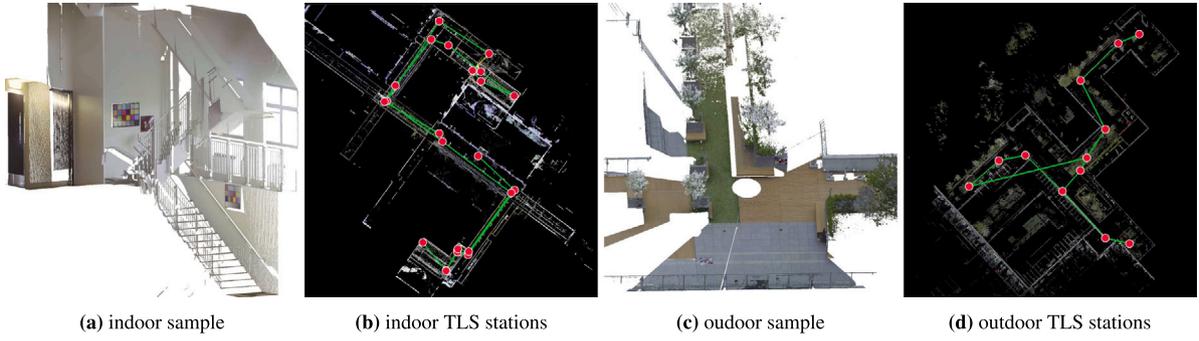

(a) indoor sample  (b) indoor TLS stations  (c) oudoor sample  (d) outdoor TLS stations

**Fig. 3.** Sample data produced by Leica RTC360. (a) a sampled point cloud in a staircase with color checkers pasted on the wall; (b) an overview of scanned floor/staircase and the registration graph, where the red circles represent the positions of scan sites and the green lines keep the geometric constraints among these sites. (c) a sampled point cloud in an outdoor environment with color checker pasted on the wall and the ground; (d) an overview of the scanned environment and the registration graph of TLS. (For interpretation of the references to color in this figure legend, the reader is referred to the web version of this article.)

**Table 2**
Details of the PolyU-BPCoMa dataset.

| Environment | Sequence | GNSS | IMU (thousand) | Panoramas | Panorama pixels (billion) | Velodyne scans | Velodyne points (million) | TLS stations | TLS points (billion) |
|---|---|---|---|---|---|---|---|---|---|
| Terrace (outdoor) | 01 | 1552 | 121.7 | 9183 | 308.1 | 6141 × 2 | 186.5 | 17 | 1.6 |
| | 02 | 1138 | 89.6 | 6725 | 225.7 | 4514 × 2 | 135.4 | | |
| Office (indoor) | 01 | 0 | 59.8 | 4417 | 148.2 | 2979 × 2 | 116.0 | 10 | 1.7 |
| | 02 | | 65.2 | 4817 | 161.6 | 3244 × 2 | 126.4 | | |
| Corridor (indoor) | 01 | 0 | 201.3 | 15048 | 504.9 | 10016 × 2 | 381.5 | 30 | 5.0 |
| | 02 | | 166.1 | 12421 | 416.8 | 8271 × 2 | 315.6 | | |
| Staircase (indoor) | 01 | 0 | 143.1 | 10699 | 359.0 | 7131 × 2 | 274.7 | 22 | 3.6 |
| | 02 | | 118.5 | 8828 | 296.2 | 5895 × 2 | 221.9 | | |
| Tunnel (indoor) | 01 | 0 | 83.6 | 6209 | 208.3 | 4162 × 2 | 162.1 | 12 | 2.0 |
| | 02 | | 67.1 | 4956 | 166.3 | 3336 × 2 | 130.2 | | |

To improve the user experience, the steps 1 and 2 in the pipeline has already been completed during dataset preparation. In addition, we provide the code for the step 3 and 7 in our open-source development kits. Therefore, data users will only need to focus on the steps 4, 5 and 6.

### 5.2. Geometrical error metric

The geometrical error is a quantitative measurement of the mobile mapping accuracy. It is defined as the average distance from a vertex point of a color checker observed in the MLS point cloud to its counterpart point observed in the TLS point cloud. This is exactly the final point-to-point loss of the ICP alignment in the benchmark pipeline. Fig. 6 demonstrates the geometrical error computation. Formally, the geometrical error $\mathcal{E}^G$ is computed by

$$\mathcal{E}^G = \frac{1}{4N} \sum_{i=1}^{N} \sum_{j=1}^{4} \left\| \mathbf{p}_j^i - \mathbf{q}_j^i \right\|_2 \tag{1}$$

where $N$ is the number of color checkers; $\mathbf{p}_j^i$ and $\mathbf{q}_j^i$, as shown in Fig. 6, are Cartesian coordinates of a pair of corresponding vertexes on a checker board. ICP is commonly used to finely align two point clouds that have small pose discrepancy. However, in this dataset, the coordinate difference of two target point sets may be too large to fall in the ICP's convergence basin. To solve this issue, we alternatively solve the least squares fitting (Arun et al., 1987) of them in the implementation. This makes the alignment result irrelevant to the initial pose difference of TLS and MLS clouds.

### 5.3. Colorization error metric

The colorization error is used to quantitatively evaluate the accuracy of point cloud colorization. Unlike the geometrical error computation which only utilizes the four vertexes of each color checker, the colorization error computation will make use of all points scanned from the color checkers. For the $i$th color checker observed from the TLS cloud and the MLS cloud, we refer to them as a reference board and a source board, respectively. The points on the reference board are denoted by $\mathcal{P}^i = \{\mathbf{p}_0, \mathbf{p}_1, \ldots, \mathbf{p}_i, \ldots\}$ while the points on the source board are denoted by $\mathcal{Q}^i = \{\mathbf{q}_0, \mathbf{q}_1, \ldots, \mathbf{q}_i, \ldots\}$. Every point has at least six properties including $(x, y, z, r, g, b)$ where the $r, g, b$ store the three-channel color information.

At first, $\mathcal{P}^i$ and $\mathcal{Q}^i$ are subsampled using a 1 cm voxel filter to ensure a uniform point density. Then, the RGB color representation is converted to the HSL color space, whereby each point will have $(x, y, z, h, s, l)$ values. Since cameras hardware and imaging settings differ between our Backpack multisensorial system and the Leica RTC 360, the saturation and lightness of images output from these two devices might have some difference. However, there will be little difference in the hue channel which prompts us to calculate the colorization error only taking into account the hue values. Formally, the colorization error $\mathcal{E}^C$ is computed by

$$\mathcal{E}^G = \frac{1}{N} \sum_{i=1}^{N} \frac{1}{M_i} \sum_{j=1}^{M_i} \left| \mathbf{p}_j^i(h) - \mathbf{q}_j^i(h) \right| \tag{2}$$

where $N$ is the number of color checkers; $M_i$ is the number of subsampled points on the $i$th color checker; $\mathbf{p}_j^i$ (h) and $\mathbf{q}_j^i$ (h) denote the hue values of the $j$th point on the $i$th reference board and source board, respectively. The correspondence of $\mathbf{p}_j^i$ and $\mathbf{q}_j^i$ are established through *nearest neighbor search*.

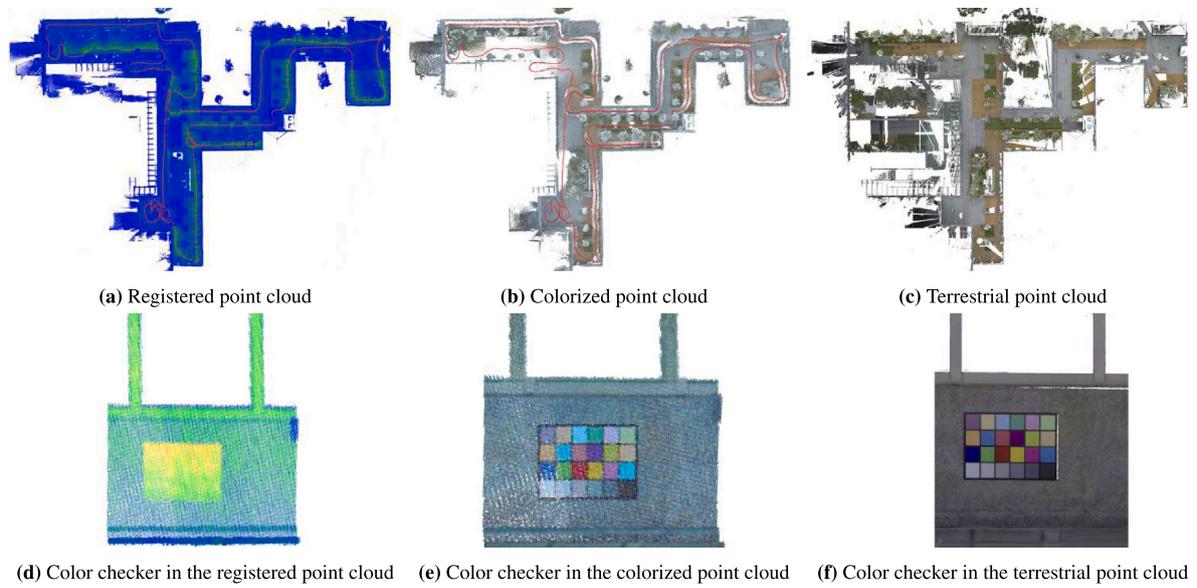

**Fig. 4.** Comparison among the registered point cloud, the colorized point cloud and the terrestrial point cloud. (a) registered point cloud generated by a SLAM method (Chen et al., 2022); (b) Point cloud colorization; (c) ground-truth data generated by TLS; (d), (e) and (f) show a color checker in the same place observed from (a), (b) and (c) respectively. Note that the registered point cloud is rendered by point intensity values for visualization while the colorized and the terrestrial point clouds are rendered by RGB values. (For interpretation of the references to color in this figure legend, the reader is referred to the web version of this article.)

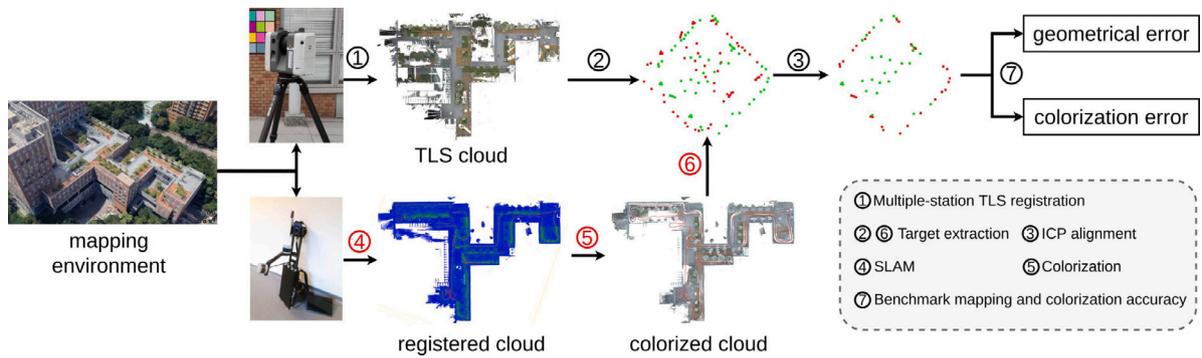

**Fig. 5.** The pipeline of benchmarking mapping and colorization accuracy. The data and code for the step 1, 2, 3, 7 are available in our open-source dataset and development kits. Hence, data users are only required to process the steps 4, 5, and 6 which are highlighted in red in the pipeline.

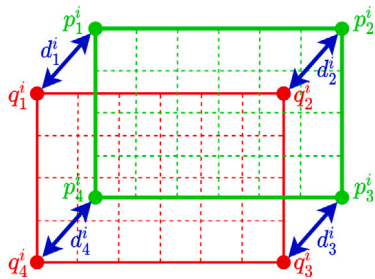

**Fig. 6.** Demonstration of geometrical error computation for a single color checker. $p_1^i, p_2^i, p_3^i$, and $p_4^i$ are the four vertex coordinates of the $i$-th color checker scanned by TLS; by contrast, $q_1^i, q_2^i, q_3^i$, and $q_4^i$ represent the same group of points but they are scanned by MLS and registered by a certain SLAM method. $d_1^i, d_2^i, d_3^i$, and $d_4^i$ represent the Euclidean distance between a pair of vertex points. (For interpretation of the references to color in this figure legend, the reader is referred to the web version of this article.)

## 6. Example dataset usages

This section presents a series of potential applications of the proposed dataset in robotics, vision and surveying communities. We provide an open-sourced development kit which includes the codes for computing geometrical and colorization errors. The kit also exhibits some quantitative results as the example and baseline of using this dataset.

### 6.1. LiDAR and vision odometry

The proposed dataset provides laser scans, panoramic RGB images and IMU readings at 20 Hz, 30 Hz and 400 Hz, respectively. These data are sufficient for IMU-aided LiDAR or vision odometry. Additionally, GNSS RTK-fixed signals are recorded at 5 Hz, and the uncertainty of a piece of GNSS measurement is recorded as a 3 × 3 covariance matrix. So the GNSS/IMU filtering results can be utilized as the global reference for LiDAR and visual odometry. Fig. 7 compares the trajectories estimated by different methods.

For benchmarking purposes, the GNSS-RTK measurements can be utilized to evaluate odometry accuracy. However, not all RTK measurements are accurate enough for evaluation due to the multipath

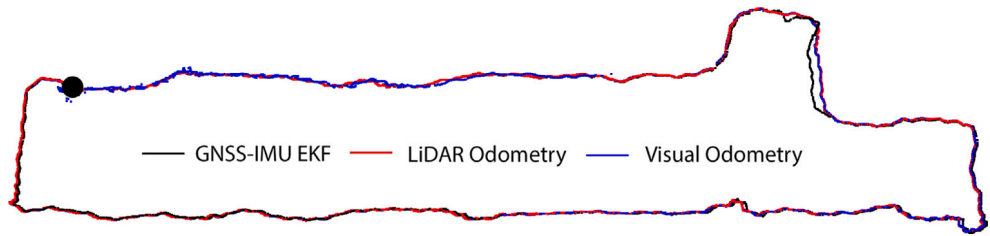

**Fig. 7.** Trajectories of *01_Terrace* estimated by GNSS/IMU EKF filtering and our LiDAR and Visual odometry methods. Black dot represents the start and stop position.

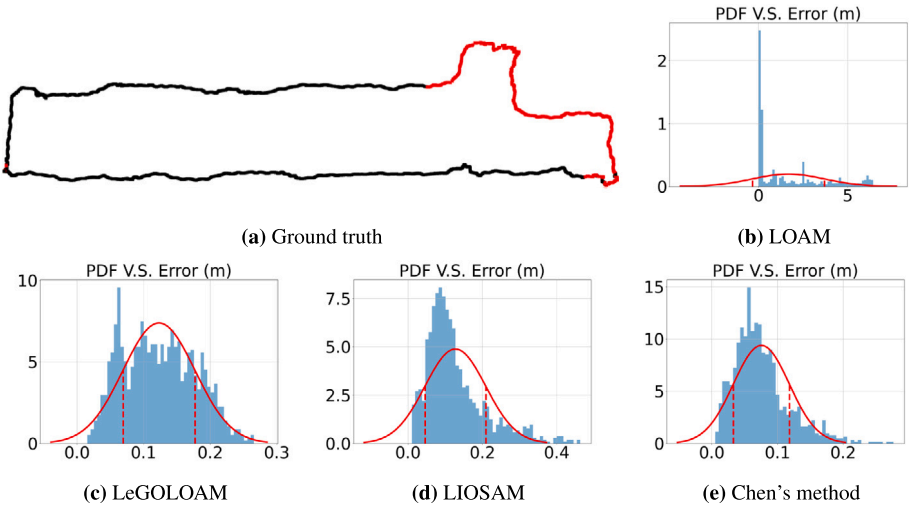

**Fig. 8.** LiDAR odometry example: (a) the ground-truth trajectory generated by GNSS-RTK measurements among which the black line segment denotes those fixed solutions, namely with the horizontal covariance less than 0.0001 and the vertical covariance less than 0.0009 while the red line means the opposite condition. (b)–(e) compare the performances of four methods. PDF denotes probability density function. (For interpretation of the references to color in this figure legend, the reader is referred to the web version of this article.)

**Table 3**
Baseline result of the odomtry task. Statistic distribution of odometry errors are studied. In the table, the check points means the number of trajectory points selected to compute errors.

| Method | Check points | $\mu$ | $\sigma$ | RMSD |
| --- | --- | --- | --- | --- |
| LOAM | 1148 | 1.680 | 2.019 | 2.627 |
| LeGOLOAM | 1148 | 0.123 | 0.054 | 0.134 |
| LIOSAM | 1148 | 0.127 | 0.082 | 0.151 |
| Chen's method | 1148 | 0.075 | 0.043 | 0.087 |

$\mu$: mean error; $\sigma$: standard deviation; RMSD: root-mean-square deviation.

effect of satellite signals. Accordingly, we have to first filter out those RTK measurements whose East/North covariance is greater than 0.0001 m$^2$ or Elevation covariance is greater than 0.0009 m$^2$. This can ensure that the remaining measurements have 1 cm and 3 cm accuracy for the East/North and Elevation directions, respectively. Taking the *01_Terrace* sequence as an example, four SLAM methods are tested and compared, including LOAM (Zhang and Singh, 2017), LeGOLOAM (Shan and Englot, 2018), LIOSAM (Shan et al., 2020) and Chen's method (Chen et al., 2021). As shown in Fig. 8, the histograms demonstrate the trajectory error distributions of different methods. Table 3 reports the statistical analysis of four methods' odometry errors.

*6.2. LiDAR colorized mapping*

We use our LiDAR SLAM algorithm detailed in Chen et al. (2022) to generate a point cloud map utilizing the proposed dataset. Fig. 9 shows an example of mobile LiDAR mapping using *01_Office* data. In this example, there are six color checkers in the scenario. Therefore, two groups of vertex points (24 in each) are extracted from TLS cloud and MLS cloud separately. These points can be used to benchmark mapping accuracy.

As the Ladybug5P camera used in the dataset directly outputs stitched panoramic images in a 2:1 aspect ratio, it is readily to convert image row-column indexes to yaw-pitch angles in a spherical coordinate system. In addition, points collected from spinning laser scanners inherently contains azimuth and *ring* information. Therefore, given the extrinsic parameters between the scanners and cameras, the colorization work becomes a 3D-to −2D spherical projection. Fig. 10 shows an example of projecting a 3D scan onto a 2D image. Fig. 11 shows the final SLAM and colorization result. This example uses the data of *01_Terrace* in the dataset and Table 4 gives the quantitative evaluation as a baseline result.

*6.3. Terrestrial cloud registration*

The PolyU-BPCoMa dataset has recorded 91 stations of terrestrial laser scanning data with ground-truth poses generated from Leica Cyclone REGISTER 360 software. After a fine registration, the pose difference among scanning sites in a dataset sequence is zero. Taking this as ground truth, users can artificially rotate or translate certain scans to test and quantitatively benchmark their terrestrial cloud registration algorithms. Users can compute the rotational error and translational error separately against the ground truth poses. Fig. 12 shows an example of aligning two TLS scans from the dataset. In this surveying scenario, 17 TLS point clouds are registered together utilizing the Leica Cyclone REGISTER 360 software. The average cloud-to-cloud overlap is 72%, and the absolute mean cloud-to-cloud error is 0.003 m. For the registration overlap and accuracy of other data sequences, readers are referred to the registration report documents in the dataset repository.

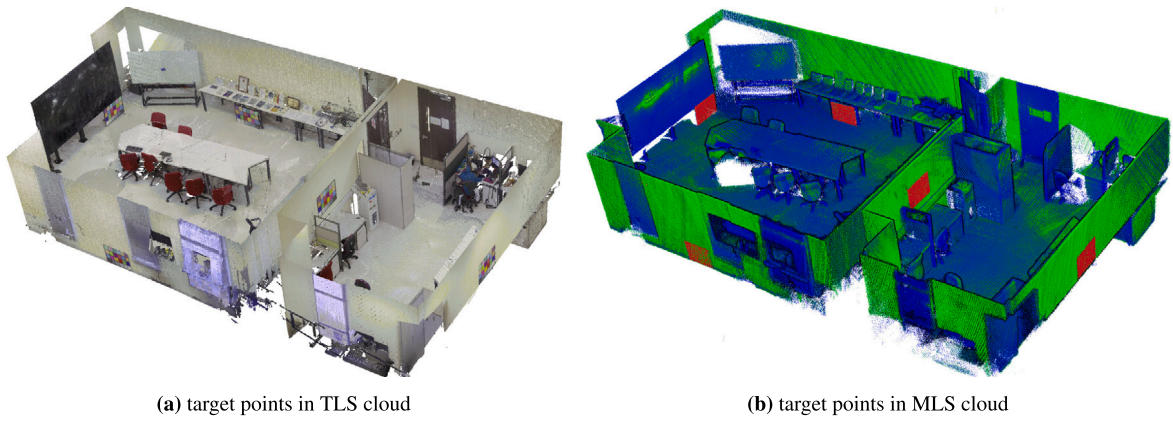

(**a**) target points in TLS cloud        (**b**) target points in MLS cloud

**Fig. 9.** LiDAR mapping example: two groups of vertex points are extracted from TLS cloud and MLS cloud separately. They are used to compute a geometrical error for benchmark purpose. In (b), the color checker boards are highlighted in red. (For interpretation of the references to color in this figure legend, the reader is referred to the web version of this article.)

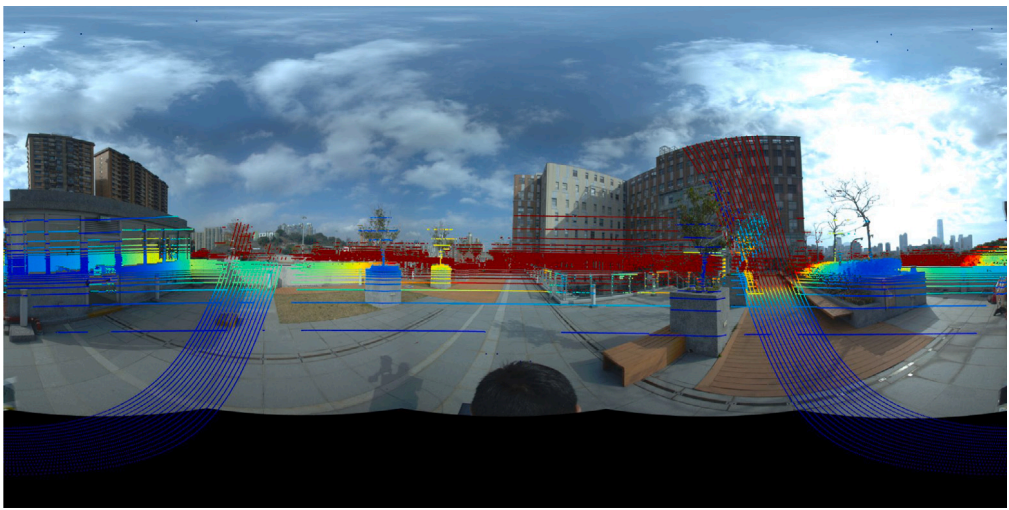

**Fig. 10.** Projecting a 3D scan onto a spherical image. The colors of scan points are rendered by their range value. The bottom black rows are Ladybug5P's blind fields of sight. (For interpretation of the references to color in this figure legend, the reader is referred to the web version of this article.)

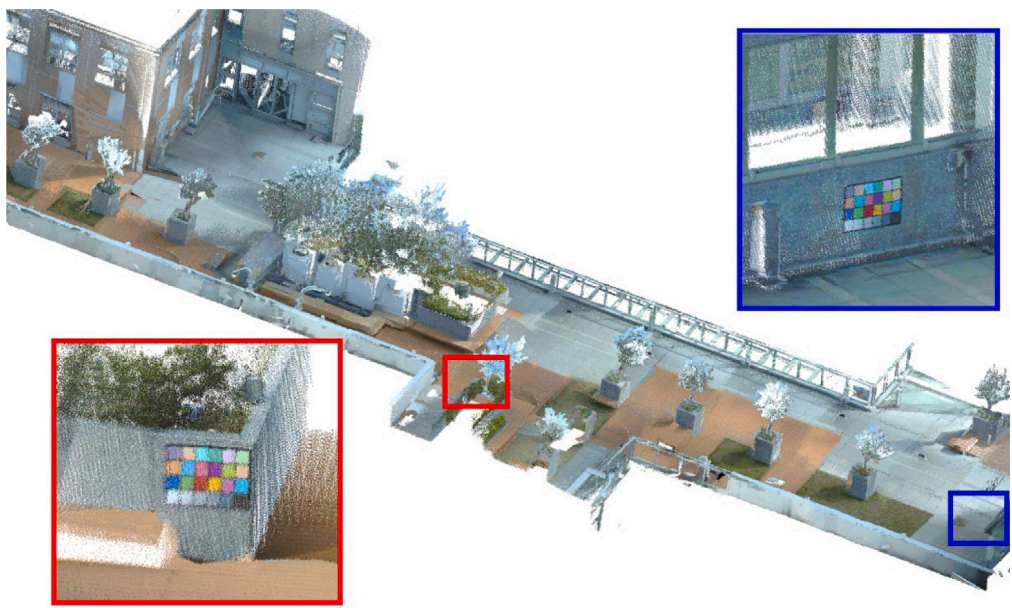

**Fig. 11.** Colorized mapping example: panoramic images are used to colorize LiDAR points after a SLAM process. Red and blue boxes show local details of color checkers in the scenario. (For interpretation of the references to color in this figure legend, the reader is referred to the web version of this article.)

**Table 4**
Baseline result: the geometrical error and colorization error of *01_Terrace*. T*x* denotes the x-*th* target board in the surveyed area. The unit of geometrical error is meter, while the unit of colorization error is hue value with the range from 0 to 1.

|  | T1 | T2 | T3 | T4 | T5 | T6 | T7 | T8 | Average |
|---|---|---|---|---|---|---|---|---|---|
| Geometrical error | 0.0891 | 0.0657 | 0.0446 | 0.0738 | 0.0756 | 0.0866 | 0.1062 | 0.0444 | 0.0733 |
| Colorization error | 0.1397 | 0.1938 | 0.1148 | 0.1582 | 0.1219 | 0.1546 | 0.1414 | 0.1204 | 0.1431 |

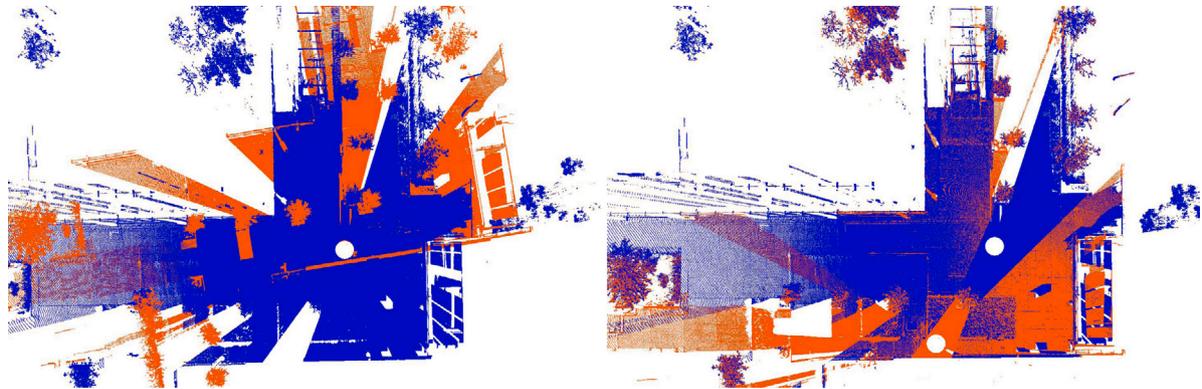

(**a**) initial pose of two TLS scans    (**b**) two TLS scans after registration

**Fig. 12.** TLS cloud registration example: the target and source TLS scans are represented in blue and orange colors respectively. (a) and (b) show the clouds before and after registration. (For interpretation of the references to color in this figure legend, the reader is referred to the web version of this article.)

*6.4. Other usages*

Taking advantage of the rich sensor resources, the proposed dataset has many other usages such as GNSS/IMU navigation, sensor fusion-based SLAM, geometrical modeling, building information modeling (BIM) generation, LiDAR or vision-based loop closure detection, etc.

## 7. Conclusion and future work

This paper documents a multisensorial dataset which is distinctively positioned towards mobile colorized mapping. This dataset incorporates sensor types of GNSS, IMU, LiDAR and spherical imaging on a backpack platform. Ground truth data are provided by Leica RTC360, a high-end terrestrial laser scanner. Color checkers are adopted as benchmark targets. They are recovered as colorized point sets by users' mapping algorithms, and then are compared against their counterparts observed from the TLS. In the benchmark pipeline, geometrical error and colorization error are evaluated simultaneously. Future work includes collecting more data sequences and long-term dataset maintenance.

**CRediT authorship contribution statement**

**Wenzhong Shi:** Supervision, Review, Funding acquisition. **Pengxin Chen:** Conceptualization, Methodology, Writing, Dataset design and capture. **Muyang Wang:** Visual odometry validation. **Sheng Bao:** GNSS-IMU EKF validation. **Haodong Xiang:** Loop closure validation. **Yue Yu:** Backpack mechanical design. **Daping Yang:** Data capture.

**Declaration of competing interest**

The authors declare that they have no known competing financial interests or personal relationships that could have appeared to influence the work reported in this paper.

**Data availability**

Data will be made available on request.

**Acknowledgments**

This work was supported by The Hong Kong Polytechnic University (CD03, 1-ZVN6); The State Bureau of Surveying and Mapping, P.R. China (1-ZVE8); and Hong Kong Research Grants Council (T22-505/19-N).

**References**

Arun, K.S., Huang, T.S., Blostein, S.D., 1987. Least-squares fitting of two 3-D point sets. IEEE Trans. Pattern Anal. Mach. Intell. (5), 698–700.

Barnes, D., Gadd, M., Murcutt, P., Newman, P., Posner, I., 2020. The Oxford Radar RobotCar dataset: A radar extension to the Oxford robotcar dataset. In: 2020 IEEE International Conference on Robotics and Automation. ICRA, IEEE, pp. 6433–6438.

Behley, J., Garbade, M., Milioto, A., Quenzel, J., Behnke, S., Stachniss, C., Gall, J., 2019. SemanticKITTI: A dataset for semantic scene understanding of LiDAR sequences. In: Proceedings of the IEEE/CVF International Conference on Computer Vision. pp. 9297–9307.

Besl, P.J., McKay, N.D., 1992. A method for registration of 3-D shapes. IEEE Trans. Pattern Anal. Mach. Intell. 14 (2), 239–256. http://dx.doi.org/10.1109/34.121791.

Burri, M., Nikolic, J., Gohl, P., Schneider, T., Rehder, J., Omari, S., Achtelik, M.W., Siegwart, R., 2016. The EuRoC micro aerial vehicle datasets. Int. J. Robot. Res. 35 (10), 1157–1163.

Caesar, H., Bankiti, V., Lang, A.H., Vora, S., Liong, V.E., Xu, Q., Krishnan, A., Pan, Y., Baldan, G., Beijbom, O., 2020. nuScenes: A multimodal dataset for autonomous driving. In: Proceedings of the IEEE/CVF Conference on Computer Vision and Pattern Recognition. pp. 11621–11631.

Carlevaris-Bianco, N., Ushani, A.K., Eustice, R.M., 2016. University of Michigan North Campus long-term vision and lidar dataset. Int. J. Robot. Res. 35 (9), 1023–1035.

Ceriani, S., Fontana, G., Giusti, A., Marzorati, D., Matteucci, M., Migliore, D., Rizzi, D., Sorrenti, D.G., Taddei, P., 2009. Rawseeds ground truth collection systems for indoor self-localization and mapping. Auton. Robots 27 (4), 353–371.

Chen, P., Luo, Z., Shi, W., 2022. Hysteretic mapping and corridor semantic modeling using mobile LiDAR systems. ISPRS J. Photogramm. Remote Sens. 186, 267–284.

Chen, P., Shi, W., Bao, S., Wang, M., Fan, W., Xiang, H., 2021. Low-drift odometry, mapping and ground segmentation using a backpack LiDAR system. IEEE Robot. Autom. Lett. 6 (4), 7285–7292. http://dx.doi.org/10.1109/LRA.2021.3097060.

Cheng, Y., Jiang, M., Zhu, J., Liu, Y., 2021. Are we ready for unmanned surface vehicles in inland waterways? The USVInland multisensor dataset and benchmark. IEEE Robot. Autom. Lett. 6 (2), 3964–3970.

Gehrung, J., Hebel, M., Arens, M., Stilla, U., 2017. An approach to extract moving objects from mls data using a volumetric background representation. ISPRS Ann. Photogramm. Remote Sens. Spatial Inform. Sci. 4.

Geiger, A., Lenz, P., Urtasun, R., 2012. Are we ready for autonomous driving? the KITTI vision benchmark suite. In: 2012 IEEE Conference on Computer Vision and Pattern Recognition. IEEE, pp. 3354–3361.

Geyer, J., Kassahun, Y., Mahmudi, M., Ricou, X., Durgesh, R., Chung, A.S., Hauswald, L., Pham, V.H., Mühlegg, M., Dorn, S., et al., 2020. A2D2: Audi autonomous driving dataset. arXiv preprint arXiv:2004.06320.

Gong, Z., Li, J., Luo, Z., Wen, C., Wang, C., Zelek, J., 2021. Mapping and semantic modeling of underground parking lots using a backpack LiDAR system. IEEE Trans. Intell. Transp. Syst. 22 (2), 734–746. http://dx.doi.org/10.1109/TITS.2019.2955734.

Hess, W., Kohler, D., Rapp, H., Andor, D., 2016. Real-time loop closure in 2D LIDAR SLAM. In: 2016 IEEE International Conference on Robotics and Automation. ICRA, IEEE, pp. 1271–1278.

Huang, A.S., Antone, M., Olson, E., Fletcher, L., Moore, D., Teller, S., Leonard, J., 2010. A high-rate, heterogeneous data set from the DARPA urban challenge. Int. J. Robot. Res. 29 (13), 1595–1601.

Julin, A., Kurkela, M., Rantanen, T., Virtanen, J.-P., Maksimainen, M., Kukko, A., Kaartinen, H., Vaaja, M.T., Hyyppä, J., Hyyppä, H., 2020. Evaluating the quality of TLS point cloud colorization. Remote Sens. 12 (17), 2748.

Maddern, W., Pascoe, G., Linegar, C., Newman, P., 2017. 1 year, 1000 km: The Oxford RobotCar dataset. Int. J. Robot. Res. 36 (1), 3–15.

Majdik, A.L., Till, C., Scaramuzza, D., 2017. The Zurich urban micro aerial vehicle dataset. Int. J. Robot. Res. 36 (3), 269–273.

Nguyen, T.-M., Yuan, S., Cao, M., Lyu, Y., Nguyen, T.H., Xie, L., 2022. NTU VIRAL: A visual-inertial-ranging-lidar dataset, from an aerial vehicle viewpoint. The International Journal of Robotics Research 41 (3), 270–280.

Pandey, G., McBride, J.R., Eustice, R.M., 2011. Ford campus vision and lidar data set. Int. J. Robot. Res. 30 (13), 1543–1552.

Pfrommer, B., Sanket, N., Daniilidis, K., Cleveland, J., 2017. Penncosyvio: A challenging visual inertial odometry benchmark. In: 2017 IEEE International Conference on Robotics and Automation. ICRA, IEEE, pp. 3847–3854.

Ramezani, M., Wang, Y., Camurri, M., Wisth, D., Mattamala, M., Fallon, M., 2020. The newer college dataset: Handheld LiDAR, inertial and vision with ground truth. In: 2020 IEEE/RSJ International Conference on Intelligent Robots and Systems. IROS, IEEE, pp. 4353–4360.

Rogers, J.G., Gregory, J.M., Fink, J., Stump, E., 2020. Test your SLAM! the SubT-Tunnel dataset and metric for mapping. In: 2020 IEEE International Conference on Robotics and Automation. ICRA, IEEE, pp. 955–961.

Schubert, D., Goll, T., Demmel, N., Usenko, V., Stückler, J., Cremers, D., 2018. The TUM VI benchmark for evaluating visual-inertial odometry. In: 2018 IEEE/RSJ International Conference on Intelligent Robots and Systems. IROS, IEEE, pp. 1680–1687.

Shan, T., Englot, B., 2018. LeGO-LOAM: Lightweight and ground-optimized LiDAR odometry and mapping on variable terrain. In: 2018 IEEE/RSJ International Conference on Intelligent Robots and Systems. IROS, IEEE, pp. 4758–4765.

Shan, T., Englot, B., Meyers, D., Wang, W., Ratti, C., Daniela, R., 2020. LIO-SAM: Tightly-coupled lidar inertial odometry via smoothing and mapping. In: IEEE/RSJ International Conference on Intelligent Robots and Systems. IROS, IEEE, pp. 5135–5142.

Smith, M., Baldwin, I., Churchill, W., Paul, R., Newman, P., 2009. The new college vision and laser data set. Int. J. Robot. Res. 28 (5), 595–599.

Wen, C., Dai, Y., Xia, Y., Lian, Y., Tan, J., Wang, C., Li, J., 2020. Toward efficient 3-D colored mapping in GPS-/GNSS-denied environments. IEEE Geosci. Remote Sens. Lett. 17 (1), 147–151.

Zhang, J., Singh, S., 2017. Low-drift and real-time LiDAR odometry and mapping. Auton. Robots 41 (2), 401–416.